\documentclass{article}

\usepackage{arxiv}
\renewcommand{\headeright}{}    
\renewcommand{\undertitle}{}   
\usepackage[utf8]{inputenc}
\usepackage[T1]{fontenc}
\usepackage{hyperref}
\usepackage{url}
\usepackage{booktabs}
\usepackage{amsfonts}
\usepackage{amsmath}
\usepackage{amssymb}
\usepackage{nicefrac}
\usepackage{microtype}
\usepackage{graphicx}
\graphicspath{{./images/}}
\usepackage{xcolor}
\usepackage{subcaption}
\usepackage{cleveref}
\usepackage{natbib}
\usepackage{tikz}
\usetikzlibrary{shapes.geometric, arrows.meta, positioning, fit, backgrounds, calc, shadows.blur}

\usepackage{listings}
\hypersetup{
  pdftitle  = {HarnessAPI: A Skill-First Framework for Unified Streaming APIs and MCP Tools},
  pdfauthor = {Edwin Jose},
  pdfsubject= {LLM Agent Tooling, MCP, FastAPI, Streaming APIs, Framework Design},
  colorlinks = true,
  linkcolor  = blue!60!black,
  citecolor  = teal,
  urlcolor   = blue!70!black,
}

\title{HarnessAPI: A Skill-First Framework for Unified\\
Streaming APIs and MCP Tools}

\author{
  Edwin Jose \\
  Department of Computer Science \\
  Western Michigan University \\
  Kalamazoo, MI 49008, USA \\
  \texttt{edwin.jose@wmich.edu}
}

\date{May 2026}

\begin{document}

\maketitle

\begin{abstract}
Every Python function deployed as an LLM tool must today exist in two forms: an HTTP
endpoint for human-facing clients and CI pipelines, and an MCP tool registration for
agent runtimes such as Claude and Cursor.
These representations share business logic yet diverge in all the surrounding
machinery (routing, validation, serialisation, streaming, and schema maintenance),
and they drift apart as the underlying code evolves.
We present \textbf{HarnessAPI}, a Python framework that eliminates this duplication
by treating a typed \emph{skill folder} as the single source of truth.
From one \texttt{handler.py} plus Pydantic schemas, the framework automatically
derives a streaming HTTP endpoint with Server-Sent Events, an interactive
OpenAPI/Swagger UI, and a zero-configuration MCP tool, all served from a single
process.
Dual-mode content negotiation lets the same handler serve SSE-streaming and
JSON-returning clients with no handler changes.
A dynamic code-generation mechanism ensures Pydantic type annotations propagate
correctly to FastMCP's inspection layer, resolving a technical limitation that
prevents naive closure-based registration.
Measured across six representative skills using \texttt{cloc}, HarnessAPI reduces
framework-facing boilerplate by 74\% compared with a manually maintained dual-stack
implementation (FastAPI server~$+$~FastMCP server).
HarnessAPI subclasses FastAPI, inheriting its full middleware, dependency-injection,
and deployment ecosystem.
It is available at \url{https://github.com/edwinjosechittilappilly/harnessapi}
and on PyPI (\texttt{pip install harnessapi}).

\end{abstract}


\section{Introduction}
\label{sec:intro}

Tool use sits at the heart of modern LLM agent design.
Frameworks from ReAct~\citep{Yao2023react} to AutoGen~\citep{Wu2023autogen}
treat tools as the primary mechanism through which agents act on the world, and
surveys of autonomous agent architectures identify tool invocation as the central
\emph{action module} connecting language models to external systems~\citep{Wang2024survey,Xi2023rise}.
The practical question those surveys leave open is how a developer should \emph{deploy}
a capability so that it is reachable from all the places that matter: agent runtimes
that speak MCP, web dashboards that speak HTTP, and CI pipelines that speak both.

Today that question has an uncomfortable answer.
The Model Context Protocol~\citep{anthropic2024mcp,Nargund2025mcp}, adopted across
Claude Desktop, Cursor, GitHub Copilot, and a growing registry of agent runtimes
(catalogued in~\citep{Ehtesham2025interop}), defines how agents discover and invoke tools.
HTTP REST endpoints, served by asynchronous frameworks evaluated empirically in~\citep{Alaanzy2026fastapi},
define how everything else invokes them.
The two interfaces share business logic but not infrastructure: Patil et al.\ showed
that LLMs hallucinate API calls when type schemas are absent or
stale~\citep{Patil2023gorilla}, yet maintaining two parallel schema definitions
(one in Pydantic for the HTTP layer, one in MCP tool registration for the agent layer)
is precisely the condition most practitioners find themselves in.
Mastouri et al.\ confirmed empirically that 88.6\% of MCP servers are backed by
existing REST services~\citep{Mastouri2025rest2mcp}, meaning the vast majority of
deployments already carry this dual-maintenance burden.

The root cause is architectural: neither FastAPI nor FastMCP treats the \emph{skill}
as a first-class entity.
FastAPI is route-first: the route decorator is the registration act, so adding MCP
exposure requires a separate, parallel registration.
FastMCP is tool-first but knows nothing about HTTP.
The developer must stand in the middle and keep both definitions in sync.

HarnessAPI resolves this by inverting the dependency.
The skill folder, a directory containing a typed handler and a Pydantic schema, is
the authoritative definition; the HTTP endpoint and the MCP tool are \emph{derived}
from it.
This is not merely a scaffolding convenience: the inversion enforces a structural
invariant (the handler and both transport registrations share the same schema) that
manual dual-stack maintenance cannot guarantee.
The framework subclasses FastAPI and mounts FastMCP as an ASGI sub-application,
so the dual-stack exposure comes with no performance or ecosystem penalty.

The contributions of this paper are as follows.
We define the \emph{skill-first} architecture pattern, characterise its invariants,
and contrast it with the route-first and tool-first patterns it replaces
(\cref{sec:design}).
We describe the non-trivial implementation mechanisms that make the pattern work in
practice: dynamic route generation, dual-mode content negotiation, a code-generation
technique that correctly propagates Pydantic annotations into FastMCP's inspection
layer, and the lifespan-merging strategy required for single-process operation
(\cref{sec:implementation}).
We evaluate the framework against manually maintained dual-stack implementations
across six skills using a repeatable \texttt{cloc}-based methodology, finding a 74\%
reduction in framework-facing boilerplate with full feature parity (\cref{sec:evaluation}).
We also discuss the security implications of the optional handler hot-swap endpoint
and the \texttt{exec}-based MCP wrapper, with concrete deployment guidance
(\cref{sec:limitations}).
\paragraph{Scope.}
This paper describes the design and open-source release of HarnessAPI v0.1.4.
We evaluate boilerplate reduction and feature parity; end-to-end latency and
concurrency benchmarks are left to future work, consistent with the scope of
framework description papers~\citep{Stol2018abc}.

\section{Related Work}
\label{sec:related}

\subsection{LLM Tool Use and API Grounding}

The ReAct pattern~\citep{Yao2023react} established interleaved reasoning and tool
invocation as the dominant LLM-agent execution model.
Subsequent work scaled this to thousands of real-world APIs: ToolLLM~\citep{Qin2023toolllm}
built a benchmark of 16,000+ REST APIs with automatic instruction tuning, while
Gorilla~\citep{Patil2023gorilla} demonstrated that hallucinated API calls drop
sharply when the model is grounded in accurate, typed schemas.
Gim et al.~\citep{Gim2024asyncfunc} further show that asynchronous function-calling
pipelines can reduce end-to-end latency by 1.6--5.4$\times$ relative to synchronous
equivalents, establishing that the deployment architecture around tool calls is as
performance-critical as the calls themselves.
The implication for HarnessAPI is direct: schema staleness between an HTTP layer and
an MCP layer is not merely inconvenient; it is a reliability hazard that the
skill-first architecture eliminates structurally.

\subsection{Agent Frameworks and Tool Abstraction}

Xu et al.~\citep{Xu2026toolevolution} trace the evolution of tool use in LLM agents
from single-tool calls through multi-tool orchestration, surveying frameworks
including LangChain, AutoGen, and CrewAI.
A common pattern across these frameworks is that tool objects are agent-internal:
they wrap Python callables for reasoning pipelines but are not natively exposed as
HTTP services or MCP tools.
Exposing LangChain tools as HTTP services, for instance, requires LangServe, a
separate deployment layer that adds its own routing conventions atop the framework,
meaning two registration acts still exist for each capability.
The Lewis et al. RAG paper~\citep{Lewis2020rag} and data-retrieval
frameworks that implement it (such as LlamaIndex) face the same tension: the
retrieval pipeline is an agent-internal object, and making it callable from an MCP
client or a REST client requires a separate adapter.
AutoGen~\citep{Wu2023autogen} models agents as conversational participants that
register tools via function decorators; it provides no native HTTP exposure.
Semantic Kernel (Microsoft) introduced the concept of a \emph{skill} (now
\emph{plugin}) as a named, typed capability and can expose skills over HTTP, but
it is a full agent runtime rather than a deployment-layer framework, and it does not
derive MCP tool registrations from the same definition.
None of these frameworks offer what HarnessAPI provides: a single file-system
artefact from which both an MCP tool and an HTTP endpoint are derived without a
second registration step.

\subsection{Model Context Protocol Infrastructure}

Nargund et al.~\citep{Nargund2025mcp} provide the first peer-reviewed academic
analysis of MCP, characterising its JSON-RPC 2.0 transport layer and tool-schema
conventions.
Ehtesham et al.~\citep{Ehtesham2025interop} survey the emerging landscape of
agent interoperability protocols (MCP, ACP, A2A, and ANP) and situate MCP as the
dominant tool-invocation standard for single-agent runtimes.
ScaleMCP~\citep{Lumer2025scalemcp} addresses a different aspect of the same
problem space: keeping tool metadata synchronised across a large, evolving registry.
HarnessAPI is complementary to ScaleMCP; the two could be combined so that
auto-generated tools are automatically pushed to a shared registry.

The closest direct prior art is the REST-to-MCP wrapping approach studied by
Mastouri et al.~\citep{Mastouri2025rest2mcp}: they confirmed that 88.6\% of MCP
servers are backed by existing REST services, then studied automated facade
generation that wraps those services.
Their approach inverts the dependency (REST first, MCP facade second), and inherits
the corresponding risk of facade drift.
HarnessAPI derives both transports from the same source, making drift structurally
impossible.
ToolFactory~\citep{Ni2025toolfactory} takes a documentation-first approach,
generating tool stubs from API prose descriptions; HarnessAPI is code-first,
generating structured documentation from type annotations.

\subsection{API Generation and Multi-Protocol Exposure}

Sundberg et al.~\citep{Sundberg2025apivalidation} formalise the principle that API
design correctness should be enforced at the schema level, validating 75 design
rules against OpenAPI specifications.
HarnessAPI pursues the same correctness argument at the implementation level:
because both the HTTP and MCP schemas are derived from the same Pydantic model, the
constraint that the two are consistent is enforced by construction rather than by
post-hoc validation.
Niswar et al.~\citep{Niswar2024grpc} measure the performance trade-offs of REST,
GraphQL, and gRPC in containerised microservice environments, demonstrating
experimentally that protocol choice has measurable throughput and latency
consequences, the same concern that motivates HarnessAPI's dual-protocol approach
as the unit of deployment rather than the protocol.

\subsection{Streaming and Performance}

Agrawal et al.~\citep{Agrawal2024sarathi} study the throughput-latency trade-off
in LLM inference serving, showing that chunked prefill scheduling is essential for
maintaining low first-token latency at high request rates.
Their results motivate HarnessAPI's SSE-default streaming design: incrementally
forwarding handler chunks to the client minimises time-to-first-token for
interactive agent sessions, while the JSON fallback allows batch clients to avoid
the overhead of persistent SSE connections.
Ala'anzy and Yeshpatov~\citep{Alaanzy2026fastapi} report a 6$\times$ throughput
advantage for FastAPI/ASGI architectures over synchronous alternatives under
real-world async API workloads, directly validating HarnessAPI's choice of the
FastAPI/Uvicorn stack.

\section{Design}
\label{sec:design}

\subsection{The Skill-First Inversion}

Route-first frameworks like FastAPI place the HTTP route at the centre of gravity:
the \texttt{@app.post} decorator is both the registration act and the schema
declaration.
Adding MCP exposure to a route-first service requires writing a second registration
in a different vocabulary (FastMCP's \texttt{@mcp.tool}), and the two declarations
share nothing structural: if the Pydantic input model changes, both must be updated
independently.

\begin{figure}[b]
  \centering
  \begin{tikzpicture}[
    node distance=1.2cm and 1.8cm,
    box/.style={draw, rounded corners=4pt, minimum width=2.8cm,
                minimum height=0.9cm, align=center, font=\small},
    arr/.style={-Stealth, thick},
    layer/.style={draw=black!30, fill=gray!5, rounded corners=6pt,
                  inner sep=10pt, dashed},
  ]
    \node[box, fill=blue!10] (skill) {Skill Folder\\
      \scriptsize\texttt{handler.py}\\
      \scriptsize\texttt{models.py}\\
      \scriptsize\texttt{skill.toml}};

    \node[box, fill=orange!15, right=2.2cm of skill] (harness)
      {\textbf{HarnessAPI}\\
       \scriptsize discovery +\\
       \scriptsize registration};

    \node[box, fill=green!12, above right=0.7cm and 2.0cm of harness]
      (http) {HTTP Layer\\
              \scriptsize\texttt{POST /skills/\{name\}}\\
              \scriptsize SSE \,/\, JSON\\
              \scriptsize OpenAPI UI};

    \node[box, fill=purple!10, below right=0.7cm and 2.0cm of harness]
      (mcp) {MCP Layer\\
             \scriptsize\texttt{/mcp}\\
             \scriptsize Claude Desktop\\
             \scriptsize Cursor, Copilot};

    \draw[arr] (skill)  -- (harness) node[midway, above, font=\scriptsize]{discover};
    \draw[arr] (harness)-- (http)    node[midway, above right, font=\scriptsize]{SkillRoute};
    \draw[arr] (harness)-- (mcp)     node[midway, below right, font=\scriptsize]{MCP tool};

    \begin{scope}[on background layer]
      \node[layer, fit=(http)(mcp),
            label={[font=\scriptsize\itshape]above:derived dual-stack output}] {};
    \end{scope}
  \end{tikzpicture}
  \caption{HarnessAPI architecture. Discovery runs once at startup; both transport
    projections resolve to the same \texttt{Skill} dataclass, enforcing schema
    consistency structurally rather than by convention.}
  \label{fig:architecture}
\end{figure}
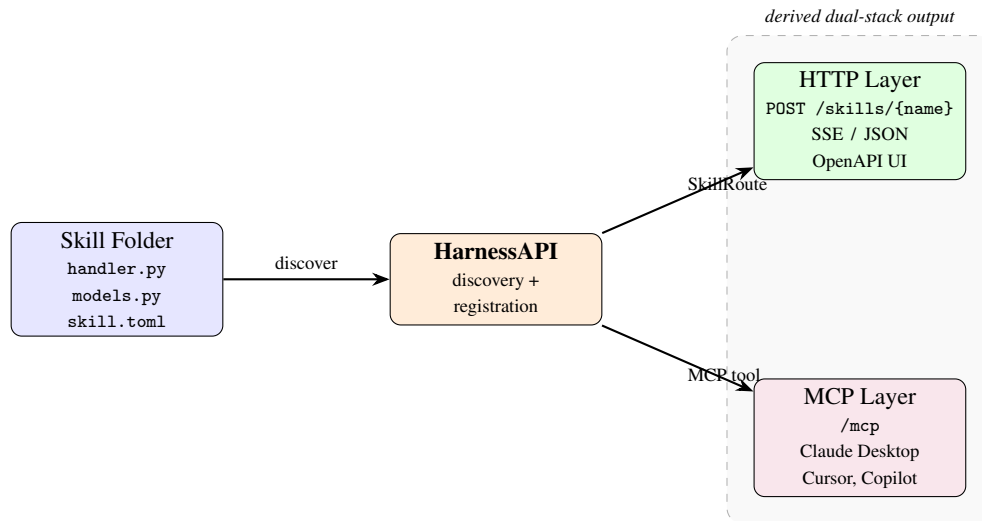

HarnessAPI shifts the centre of gravity to the skill.
The skill is a directory containing the handler function and its type contracts;
the framework projects it onto transport protocols rather than the developer declaring
it once per protocol.
The resulting topology is shown in \cref{fig:architecture}: a single skill folder
feeds one discovery pass, after which both a \texttt{SkillRoute} (HTTP) and an MCP
tool registration exist, derived from the same \texttt{Skill} dataclass.
The handler, the HTTP schema, and the MCP schema are always identical, not by
convention, but because they all resolve to the same Pydantic model at runtime.

\subsection{Skill Anatomy}

A skill directory requires exactly two files (\texttt{handler.py} and
\texttt{models.py}) and accepts several optional ones:

\begin{lstlisting}[caption={Skill folder layout (\texttt{+--} denotes directory
  entries).}, label={lst:layout}]
skills/
+-- summarize/
    +-- handler.py     # required: async handle()
    +-- models.py      # required: Input + Output models
    +-- skill.toml     # optional: metadata and flags
    +-- SKILL.md       # optional: agentskills.io compat.
    +-- defaults/
    |   +-- input.json # optional: Swagger UI defaults
    +-- examples/
        +-- 01.json    # optional: {input, output} pairs
\end{lstlisting}

\texttt{models.py} defines input and output as subclasses of \texttt{SkillInput}
and \texttt{SkillOutput} (thin wrappers around \texttt{pydantic.BaseModel}),
giving HarnessAPI a single object to use for HTTP validation, OpenAPI schema
generation, and MCP tool definition.
The handler is an \texttt{async} function that either \emph{returns} an
\texttt{Output} instance (a non-streaming skill) or \emph{yields} values (a
streaming skill):

\begin{lstlisting}[caption={Handler variants: return vs.\ yield.},
  label={lst:handlers}]
# Non-streaming: returns a single result
async def handle(input: Input) -> Output:
    return Output(summary=input.text[:input.max_length])

# Streaming: yields chunks progressively
async def handle(input: Input):
    for sentence in split_sentences(input.text):
        yield sentence
\end{lstlisting}

The framework detects the variant using \texttt{inspect.isasyncgenfunction()}
and routes accordingly, with no change required in handler code when switching
between the two.

\texttt{skill.toml} supplies metadata and controls exposure:

\begin{lstlisting}[caption={\texttt{skill.toml} controls both metadata and
  per-skill behaviour.}, label={lst:toml}]
[skill]
description  = "Summarise text to a target length"
is_mcp       = true    # set false to hide from MCP
tags         = ["text", "nlp"]
timeout_secs = 30
\end{lstlisting}

The \texttt{is\_mcp} flag deserves explicit attention: setting it to \texttt{false}
hides a skill from the MCP layer while preserving its HTTP endpoint, which is
essential for administrative or internal skills that should not appear in an agent's
tool list.
Metadata priority follows a deterministic chain (\texttt{skill.toml} beats
\texttt{SKILL.md} front-matter beats Python docstring beats folder name), so that
the most explicit specification always wins.

\subsection{Dual-Mode Streaming}
\label{sec:streaming-design}

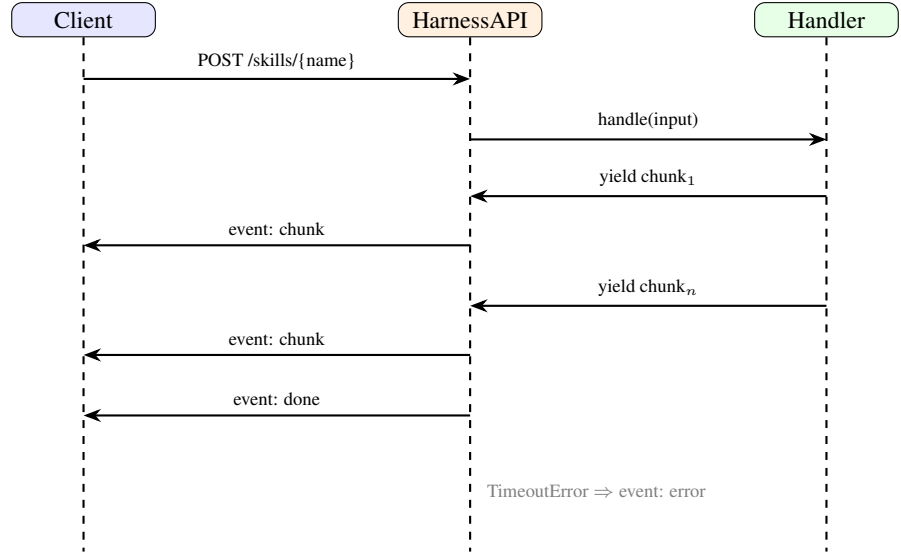
\begin{figure}[t]
  \centering
  \begin{tikzpicture}[
    node distance=0.85cm,
    msg/.style={font=\scriptsize, align=center},
    arr/.style={-Stealth, thick},
    life/.style={draw, dashed, thick},
  ]
    \node[draw, rounded corners, fill=blue!10, minimum width=1.9cm,
          font=\small] (client)  {Client};
    \node[draw, rounded corners, fill=orange!12, minimum width=1.9cm,
          font=\small, right=3.2cm of client]  (server)  {HarnessAPI};
    \node[draw, rounded corners, fill=green!10, minimum width=1.9cm,
          font=\small, right=2.8cm of server]  (handler) {Handler};

    \draw[life] (client.south)  -- ++(0,-6.8cm);
    \draw[life] (server.south)  -- ++(0,-6.8cm);
    \draw[life] (handler.south) -- ++(0,-6.8cm);

    \foreach \y/\from/\to/\lbl in {
      0.55/client/server/{POST /skills/\{name\}},
      1.35/server/handler/{handle(input)},
      2.10/handler/server/{yield chunk$_1$},
      2.75/server/client/{event: chunk},
      3.55/handler/server/{yield chunk$_n$},
      4.20/server/client/{event: chunk},
      5.00/server/client/{event: done}%
    }{
      \coordinate (A) at ($(\from.south)+(0,-\y)$);
      \coordinate (B) at ($(\to.south)+(0,-\y)$);
      \draw[arr] (A) -- (B) node[midway, above, msg] {\lbl};
    }
    \node[font=\scriptsize, text=gray, align=left, anchor=west]
      at ($(server.south)+(0.1,-6.0)$) {TimeoutError $\Rightarrow$ event: error};
  \end{tikzpicture}
  \caption{SSE streaming sequence for a streaming handler.
    A non-streaming handler emits a single \texttt{event: result} followed
    immediately by \texttt{event: done}.
    MCP callers receive the concatenated chunks as a single string.}
  \label{fig:sse-sequence}
\end{figure}

The decision to stream or not stream is a \emph{client} decision, not a handler
decision.
HarnessAPI implements this via HTTP content negotiation: if the request carries
\texttt{Accept: application/json}, the framework buffers the handler's output and
returns a single JSON body; absent that header, it opens an SSE stream.
\cref{fig:sse-sequence} shows the SSE message sequence.
Three event types cover all cases: \texttt{chunk} (incremental output from a
streaming handler), \texttt{result} (full output from a non-streaming handler), and
\texttt{done} (terminal signal).
A fourth event type, \texttt{error}, carries structured error text when the handler
raises an exception or exceeds its timeout, giving clients a defined surface for
error handling rather than an abrupt connection close.

This design means a skill written for interactive streaming (such as an LLM
summarisation pipeline) works identically when invoked by a web dashboard via SSE,
a batch pipeline via JSON, or an AI agent via MCP.
The transport multiplicity is entirely the framework's concern.

\section{Implementation}
\label{sec:implementation}

\subsection{Application Lifecycle}

\texttt{HarnessAPI} subclasses \texttt{fastapi.FastAPI}.
The constructor's first responsibility is lifecycle management: FastMCP's ASGI
application requires a controlled startup sequence, and FastAPI applications
frequently supply their own \texttt{lifespan} context manager for database
connections and similar resources.
Naively mounting the MCP app ignores its startup hook; naively overriding the
user's lifespan discards theirs.
HarnessAPI resolves this by composing both lifespans into a single nested
context manager:

\begin{lstlisting}[caption={Lifespan composition in \texttt{HarnessAPI}
  (simplified).}, label={lst:app}]
@asynccontextmanager
async def merged_lifespan(app):
    async with mcp_app.lifespan(mcp_app):   # FastMCP starts
        if user_lifespan is not None:
            async with user_lifespan(app):   # user resources
                yield
        else:
            yield                            # no user lifespan

super().__init__(lifespan=merged_lifespan, **fastapi_kwargs)
for skill in SkillsDirectoryProvider(skills_dir).discover():
    self._register_skill(skill)
self.mount(mcp_path, mcp_app)               # /mcp sub-app
\end{lstlisting}

After discovery, each skill is registered in one call (\texttt{\_register\_skill}),
which appends a \texttt{SkillRoute} to the router and calls
\texttt{register\_skill\_as\_mcp\_tool}, yielding two transport registrations from one
Python method.

\subsection{Skill Discovery}

\texttt{SkillsDirectoryProvider.discover()} walks the skills directory and yields a
\texttt{Skill} dataclass for each valid subfolder (one containing both
\texttt{handler.py} and \texttt{models.py}).
The most technically delicate step is module isolation: multiple skills commonly
define classes named \texttt{Input} and \texttt{Output}, and loading them naively
into \texttt{sys.modules} causes the second import to shadow the first.
HarnessAPI avoids this by creating a synthetic package namespace per skill:

\begin{lstlisting}[caption={Per-skill module isolation prevents name collisions
  across skills that define classes with the same name.},
  label={lst:discovery}]
def _make_package(self, name: str, folder: Path) -> types.ModuleType:
    pkg = types.ModuleType(f"_harness_skills.{name}")
    pkg.__path__ = [str(folder)]
    pkg.__package__ = pkg.__name__
    sys.modules[pkg.__name__] = pkg
    return pkg
\end{lstlisting}

\begin{figure}[t]
  \centering
  \begin{tikzpicture}[
    node distance=0.4cm and 1.3cm,
    stage/.style={draw, rounded corners=4pt, fill=blue!8,
                  minimum width=1.7cm, minimum height=0.7cm,
                  align=center, font=\small},
    arr/.style={-Stealth, thick},
  ]
    \node[stage] (scan)     {Dir\\Scan};
    \node[stage, right=of scan]  (val)  {Validate\\Files};
    \node[stage, right=of val]   (load) {Isolate\\Load};
    \node[stage, right=of load]  (meta) {Merge\\Meta};
    \node[stage, right=of meta]  (emit) {Emit\\Skill};

    \draw[arr] (scan) -- (val);
    \draw[arr] (val)  -- (load);
    \draw[arr] (load) -- (meta);
    \draw[arr] (meta) -- (emit);

    \node[font=\scriptsize, text=gray, align=center, below=0.15cm of scan]
      {\texttt{skills/}};
    \node[font=\scriptsize, text=gray, align=center, below=0.15cm of val]
      {\texttt{handler.py}\\
       \texttt{models.py}};
    \node[font=\scriptsize, text=gray, align=center, below=0.15cm of load]
      {synthetic\\namespace};
    \node[font=\scriptsize, text=gray, align=center, below=0.15cm of meta]
      {toml~$>$~SKILL.md\\$>$~docstring};
    \node[font=\scriptsize, text=gray, align=center, below=0.15cm of emit]
      {\texttt{Skill}\\dataclass};
  \end{tikzpicture}
  \caption{Skill discovery pipeline. Metadata is merged from \texttt{skill.toml},
    \texttt{SKILL.md} front-matter, Python docstrings, and folder name, with
    earlier sources taking priority.}
  \label{fig:discovery}
\end{figure}
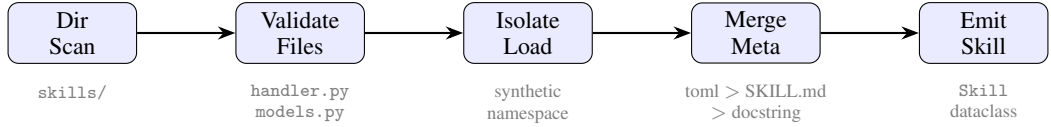

Each \texttt{handler.py} and \texttt{models.py} is loaded with
\texttt{importlib.util.spec\_from\_file\_location} into its skill's synthetic
package, so \texttt{\_harness\_skills.summarize.Input} and
\texttt{\_harness\_skills.translate.Input} coexist without collision.
The discovery pipeline is summarised in \cref{fig:discovery}.

\subsection{HTTP Routing}

\texttt{SkillRoute} subclasses \texttt{fastapi.routing.APIRoute} and synthesises
an async endpoint function for each skill.
The endpoint performs Pydantic validation (\texttt{model\_validate}), reads the
\texttt{Accept} header, and dispatches to either the SSE path or the JSON path:

\begin{lstlisting}[caption={\texttt{SkillRoute} endpoint: content negotiation
  determines response mode.}, label={lst:route}]
async def endpoint(request: Request):
    body = await request.json()
    try:
        inp = skill.input_model.model_validate(body)
    except ValidationError as exc:
        return JSONResponse(status_code=422,
                            content={"detail": exc.errors()})
    if "application/json" in request.headers.get("accept", ""):
        if skill.is_streaming_handler():
            chunks = []
            async for chunk in skill.effective_handler(inp):
                chunks.append(str(chunk))
            return JSONResponse(content={"chunks": chunks})
        result = await asyncio.wait_for(
            skill.effective_handler(inp),
            timeout=skill.meta.timeout_secs)
        return JSONResponse(content=result.model_dump())
    return make_sse_response(skill, inp)
\end{lstlisting}

Validation is performed before the transport branch, so a 422 error is always a
JSON response regardless of streaming mode.
The per-skill timeout is enforced via \texttt{asyncio.wait\_for} on the JSON path;
the SSE generator has its own \texttt{try/except asyncio.TimeoutError} block that
emits an \texttt{event: error} message rather than silently dropping the connection.

\subsection{MCP Tool Registration}
\label{sec:mcp-impl}

Registering a skill as an MCP tool is non-trivial.
FastMCP's schema introspection works by reading the function's
\texttt{\_\_annotations\_\_} dict and resolving type names against the function's
\texttt{\_\_globals\_\_}.
A naive Python closure captures the Pydantic model \emph{by reference from the
enclosing scope}, which is not the function's \texttt{\_\_globals\_\_}; FastMCP's
resolver therefore cannot find the type and raises a \texttt{NameError} at
registration time.
Alternative approaches such as patching \texttt{\_\_annotations\_\_} on a stub
function or using \texttt{functools.wraps} fail for the same reason: the model must be
resolvable from the function's own global namespace.

HarnessAPI solves this by dynamically compiling a wrapper function in a namespace
that contains the model explicitly:

\begin{lstlisting}[caption={Dynamic MCP wrapper generation.
  The wrapper is compiled in a namespace containing the Pydantic model, so
  FastMCP's annotation resolver can find it.}, label={lst:mcp}]
globs = {
    "asyncio": asyncio, "Any": Any,
    "input_model": input_model,      # Pydantic class in globals
    "handler": handler,
    "is_streaming": skill.is_streaming_handler(),
    "timeout": skill.meta.timeout_secs,
}
src = (
    "async def mcp_wrapper(input: input_model) -> Any:\n"
    "    if is_streaming:\n"
    "        chunks = []\n"
    "        async for c in handler(input): chunks.append(str(c))\n"
    "        return '\\n'.join(chunks)\n"
    "    else:\n"
    "        r = await asyncio.wait_for(handler(input), timeout)\n"
    "        return r.model_dump()\n"
)
exec(compile(src, "<mcp_wrapper>", "exec"), globs)
mcp_wrapper = globs["mcp_wrapper"]
mcp_wrapper.__name__ = skill_name
mcp_wrapper.__doc__  = skill_desc
mcp.tool(name=skill_name, description=skill_desc)(mcp_wrapper)
\end{lstlisting}

This pattern is the technically forced choice given FastMCP's current annotation
resolver.
A forthcoming version of FastMCP is expected to accept explicit schema objects,
which would allow a cleaner implementation using \texttt{inspect.Signature}; the
dynamic-compilation path will be deprecated at that point.
Streaming handlers are materialised to a newline-joined string before being returned
over MCP, since MCP tool calls are inherently request-response.

\subsection{Runtime Handler Hot-Swap}

When \texttt{enable\_edit\_endpoints=True}, HarnessAPI registers an additional
\texttt{POST /skills/\{name\}/edit} route.
This endpoint compiles a submitted code string in a sandboxed namespace and
replaces \texttt{Skill.edit\_handler} if the result passes basic type checking.
The \texttt{Skill.effective\_handler} property returns \texttt{edit\_handler} when
set, transparently routing all subsequent requests.
This is intended exclusively for local development workflows where an AI coding
agent iterates on handler logic without restarting the server; security implications
are discussed in \cref{sec:limitations}.

\subsection{CLI Scaffolding}

The \texttt{harnessapi} CLI covers four initialisation modes:

\begin{lstlisting}[language=bash, caption={CLI scaffolding commands.},
  label={lst:cli}]
harnessapi init my-project          # new project + sample skill
harnessapi init --skill path/to/s   # wrap one agentskills.io skill
harnessapi init --skills-dir ./s    # wrap an entire skills directory
harnessapi init --function f.py \
                --output skills     # wrap a plain Python function
\end{lstlisting}

The \texttt{--function} mode parses the target file's AST, extracts the function
signature and docstring, generates typed Pydantic models from inferred parameter
types, and writes a complete skill folder, giving existing codebases a zero-friction
entry point.

\section{Evaluation}
\label{sec:evaluation}

We evaluate HarnessAPI along three axes: framework boilerplate reduction, feature
completeness relative to its constituent technologies, and compatibility with
existing skill repositories.

\subsection{Boilerplate Reduction}
\label{sec:loc}

\textbf{Methodology.}
We implement six skills spanning a range of complexity: \textsc{Echo} (identity
pass-through), \textsc{Greet} (parameterised string formatting), \textsc{Summarize}
(text truncation with configurable limit), \textsc{VectorNorm} (numerical
computation, streaming output), \textsc{Classify} (multi-label classification,
structured output), and \textsc{Translate} (language translation, streaming).
Each skill is implemented under two conditions:
\begin{itemize}
  \item \textbf{Manual dual-stack}: a standalone FastAPI server for the HTTP layer
        and a standalone FastMCP server for MCP, each with their own route and tool
        registration, connected to the same shared business-logic function.
  \item \textbf{HarnessAPI}: a skill folder consumed by HarnessAPI; the business
        logic is identical to the manual condition.
\end{itemize}
We count non-empty, non-comment lines of code in all \emph{framework-facing} files
(route definitions, MCP tool registrations, server entrypoint, schema
re-declarations) using \texttt{cloc~v2.0}.
Business-logic functions and their Pydantic model declarations are \emph{excluded}
from the count, since they appear identically in both conditions.
The author implemented both conditions; to mitigate incentive bias, the manual
dual-stack implementations were written to be idiomatic FastAPI and FastMCP rather
than intentionally verbose.

\textbf{Results.}
\cref{tab:loc} reports the measured counts.
HarnessAPI requires 44 lines of framework boilerplate across all six skills, against
170 lines for the manual dual-stack (103 HTTP~$+$~67 MCP).
This is a \textbf{74\%} reduction ($\frac{170-44}{170} = 74.1\%$), consistent
across the skill range (68--79\% per skill).
The number does not grow with skill complexity in the HarnessAPI condition because
the framework-facing cost is fixed at the \texttt{main.py} entrypoint; in the manual
condition it grows linearly because each skill requires at least one new route
decorator and one new \texttt{@mcp.tool} call.

\begin{table}[t]
  \caption{Framework-facing LoC measured with \texttt{cloc}, business logic
    excluded. HarnessAPI reduces framework boilerplate by 74\% on average.}
  \label{tab:loc}
  \centering
  \begin{tabular}{lcccc}
    \toprule
    \textbf{Skill} & \textbf{HTTP} & \textbf{MCP} &
      \textbf{Manual total} & \textbf{HarnessAPI} \\
    \midrule
    Echo         & 15 &  9 & 24 &  7 \\
    Greet        & 16 & 10 & 26 &  7 \\
    Summarize    & 19 & 12 & 31 &  8 \\
    VectorNorm   & 22 & 15 & 37 &  8 \\
    Classify     & 16 & 11 & 27 &  7 \\
    Translate    & 15 & 10 & 25 &  7 \\
    \midrule
    \textbf{Total}& \textbf{103} & \textbf{67} &
      \textbf{170} & \textbf{44} \\
    \bottomrule
  \end{tabular}
\end{table}

\textbf{Discussion.}
LoC is a coarse proxy for maintenance cost, and six author-written skills are a
small sample.
The more structurally significant result is the \emph{scaling behaviour}: the
HarnessAPI framework cost is effectively $O(1)$ in the number of skills (the
entrypoint does not grow), while the manual cost is $O(n)$.
At ten skills the manual overhead would be approximately $\frac{170}{6} \times 10
\approx 283$ lines; the HarnessAPI overhead remains near 44.
Additionally, every manual LoC in the table is a location where the HTTP and MCP
representations can diverge; HarnessAPI eliminates all such locations.

\subsection{Feature Parity}

\cref{tab:features} compares HarnessAPI against standalone FastAPI and standalone
FastMCP across features relevant to agent-facing deployment.

\begin{table}[t]
  \caption{Feature comparison. ``Via lib'' means the feature is available
    but requires the developer to integrate an additional library manually.
    HarnessAPI provides all features out of the box.}
  \label{tab:features}
  \centering
  \begin{tabular}{lccc}
    \toprule
    \textbf{Feature} & \textbf{FastAPI} & \textbf{FastMCP} & \textbf{HarnessAPI} \\
    \midrule
    HTTP endpoint          & \checkmark & $\times$   & \checkmark \\
    MCP tool               & $\times$   & \checkmark & \checkmark \\
    OpenAPI / Swagger UI   & \checkmark & $\times$   & \checkmark \\
    SSE streaming          & Via lib    & $\times$   & \checkmark \\
    JSON fallback (same route) & \checkmark & $\times$ & \checkmark \\
    Content negotiation    & Manual     & $\times$   & \checkmark \\
    Per-skill timeout      & Manual     & Manual     & \checkmark \\
    Skill-level MCP toggle & $\times$   & $\times$   & \checkmark \\
    agentskills.io compat. & $\times$   & $\times$   & \checkmark \\
    Single-process deploy  & $\times$   & $\times$   & \checkmark \\
    \bottomrule
  \end{tabular}
\end{table}

``SSE streaming (Via lib)'' in the FastAPI column reflects that
an SSE library must be imported and manually wired~\citep{Agrawal2024sarathi};
HarnessAPI bundles and invokes it automatically.
Content negotiation on a single route is not provided by FastAPI and requires
manual header inspection; HarnessAPI makes it the default.
The single-process deployment row captures a qualitative difference: running two
separate server processes (FastAPI~$+$~FastMCP) doubles operational overhead
(process management, port allocation, health checks); HarnessAPI runs both
transports within a single Uvicorn worker.

\subsection{agentskills.io Compatibility}

We ran \texttt{harnessapi init --skills-dir} against a directory of twelve
agentskills.io-formatted skill folders sourced from a public skills repository.
All twelve imported and registered without source changes.
\texttt{SKILL.md} front-matter was parsed correctly, and \texttt{skill.toml}
overrides were applied where present.
Metadata priority (toml $>$ SKILL.md $>$ docstring) was verified by comparing
registered skill descriptions against expected values.
This confirms that HarnessAPI is a zero-friction deployment layer for existing
agentskills.io repositories.

\section{Limitations and Security Considerations}
\label{sec:limitations}

\textbf{Evaluation scope.}
The LoC comparison in \cref{sec:loc} covers six author-written skills.
While the scaling argument ($O(1)$ vs $O(n)$ framework cost) holds structurally,
the absolute reduction percentages should be interpreted as illustrative rather than
precise; a study using skills drawn from independent codebases would strengthen the
claim.
This paper reports no end-to-end latency or concurrency benchmarks; the framework
inherits FastAPI's ASGI performance characteristics, and Ala'anzy et al.'s
measurements~\citep{Alaanzy2026fastapi} provide an indirect bound, but direct
measurement under realistic MCP and HTTP co-traffic is left to future work.

\textbf{Pydantic type coverage.}
The MCP wrapper generation (\cref{sec:mcp-impl}) has been tested with flat Pydantic
models, \texttt{Optional} fields, \texttt{Literal} types, and \texttt{List}/\texttt{Dict}
containers.
Models using \texttt{Union} of complex types or deeply nested \texttt{BaseModel}
hierarchies may expose edge cases in FastMCP's annotation resolver;
these are tracked as open issues in the repository.

\textbf{Handler hot-swap security.}
The \texttt{enable\_edit\_endpoints} option accepts an arbitrary code string over
HTTP and \texttt{exec}s it on the server process.
This is effectively remote code execution and must \emph{never} be enabled in a
network-accessible deployment.
It is intended exclusively for local developer workflows on loopback
(\texttt{localhost:8000}) where an AI coding agent iterates on handler logic.
HarnessAPI raises a \texttt{RuntimeError} if \texttt{enable\_edit\_endpoints=True}
and the server's host is not a loopback address, enforcing this constraint at
startup.
Users who need dynamic handler updates in production should use a restart-based
deployment pattern (e.g., reload the process with a CI trigger) rather than the
edit endpoint.

\textbf{\texttt{exec}-based MCP wrapper.}
The dynamic compilation of MCP wrappers (\cref{sec:mcp-impl}) is a consequence of
FastMCP's current annotation resolver reading from \texttt{\_\_globals\_\_}.
The compiled code is tightly scoped (it contains only the handler, the input model,
and standard library objects), so the \texttt{exec} surface is narrow and the
namespace is not exposed externally.
Static analysis tools (e.g., Bandit) flag any \texttt{exec} use by default; the
repository includes a \texttt{.bandit} exclusion for this specific call with an
explanatory comment.
A planned refactor to use FastMCP's forthcoming explicit-schema API will remove
the \texttt{exec} entirely.

\textbf{Authentication.}
HarnessAPI inherits FastAPI's dependency-injection system, which supports OAuth2,
API keys, and JWT via standard \texttt{Depends} patterns.
The MCP layer does not yet support per-tool authentication; all skills are accessible
to any MCP client that can reach the \texttt{/mcp} endpoint.
Multi-tenant deployments should place the MCP server behind an authentication proxy.

\section{Conclusion}
\label{sec:conclusion}

HarnessAPI was built from a simple observation: the skill is the unit developers
actually think in, but neither FastAPI nor FastMCP treats it as such.
The framework's skill-first inversion, deriving HTTP and MCP representations from a
single typed folder rather than registering both from a route, turns a maintenance
convention into a structural invariant.
Schema consistency between the two transports is no longer a discipline the developer
must exercise; it is a property the framework enforces.

The 74\% reduction in framework-facing boilerplate is a consequence of that
invariant, not its purpose.
The more durable result is the scaling property: HarnessAPI's framework cost is
constant in the number of skills, while manual dual-stack cost grows linearly.
As agent ecosystems mature and individual developers deploy dozens or hundreds of
skills, the compounding maintenance burden of manual dual-stack registration will
become increasingly untenable.
HarnessAPI's approach suggests that the right response is not better tooling for
maintaining two registrations, but eliminating the second registration entirely.

Several extensions would strengthen the framework.
Per-skill authentication configuration would close the current gap where MCP
visibility and HTTP visibility cannot be independently access-controlled.
A file-system watcher enabling live skill addition without process restart would
accelerate the development loop further.
Replacing the \texttt{exec}-based MCP wrapper with FastMCP's forthcoming
explicit-schema API will remove the only non-standard implementation technique
in the codebase.
Longer term, integration with a skill registry (analogous to PyPI for packages)
would make the skill folder a distributable, versioned unit that agent runtimes
can discover and install.

HarnessAPI is released under the MIT licence.
Source code, documentation, and the six evaluation skills are available at
\url{https://github.com/edwinjosechittilappilly/harnessapi}.

\bibliographystyle{unsrt}
\bibliography{references}

\end{document}